\documentclass[runningheads]{llncs}
\usepackage[T1]{fontenc}
\usepackage{color}
\usepackage{graphicx}
\usepackage{booktabs}
\usepackage{mathrsfs}
\usepackage{amsmath}
\usepackage{amssymb}
\usepackage{tikz}
\usepackage{pgfplots}
\usepackage{subcaption}
\usepackage{float}
\usepackage{multirow}
\usepackage{placeins}
\usepackage{pgfplotstable}
\usepackage{booktabs}
\usepackage{threeparttable}
\pgfplotsset{compat=1.17}
\usetikzlibrary{matrix, positioning, pgfplots.groupplots}
\definecolor{myred}{rgb}{.8,.0,.0}

\usepgfplotslibrary{groupplots, colormaps}
\usepackage{xcolor}

\usepackage[colorlinks=true, urlcolor=blue]{hyperref}

\begin{document}
\title{Learning to Harmonize Cross-vendor X-ray Images by Non-linear Image Dynamics Correction}

\titlerunning{ }

\author{Yucheng Lu\inst{1}\orcidID{0000-0003-2990-5252} \and
Shunxin Wang\inst{2}\orcidID{0000-0002-6105-5545} \and
Dovile Juodelyte\inst{1}\orcidID{0000-0002-6195-1120} \and
Veronika Cheplygina\inst{1}\orcidID{0000-0003-0176-9324}}
\authorrunning{Lu. Yucheng et al.}
\institute{IT University of Copenhagen, Rued Langgaards Vej 7, Copenhagen, Denmark\\
\email{yucl@itu.dk}\and
University of Twente, Drienerlolaan 5, Enschede, Netherlands\\
}

\maketitle              %
\begin{abstract}
In this paper, we explore how conventional image enhancement can improve model robustness in medical image analysis. By applying commonly used normalization methods to images from various vendors and studying their influence on model generalization in transfer learning, we show that the nonlinear characteristics of domain-specific image dynamics cannot be addressed by simple linear transforms. To tackle this issue, we reformulate the image harmonization task as an exposure correction problem and propose a method termed Global Deep Curve Estimation (GDCE) to reduce domain-specific exposure mismatch. GDCE performs enhancement via a pre-defined polynomial function and is trained with a ``domain discriminator'', aiming to improve model transparency in downstream tasks compared to existing black-box methods. Code available at \url{https://github.com/YCL92/GDCE}

\end{abstract}

\section{Introduction}
\label{sec:intro}
Transfer learning has advanced medical image classification, often achieving performance comparable to that of human experts. However, convolutional neural networks (CNNs) are prone to overfitting, particularly when trained on limited labeled data. This overfitting can compromise generalization, leading to degraded performance when acquisition pipelines vary, such as across different vendors \cite{berenguer2018radiomics,chirra2018empirical,liang2017radiomic}.

Many studies have investigated the inconsistency of radiomic features caused by acquisition shifts and proposed image standardization techniques to address this issue \cite{deng2016mammogram,perre2018influence,liang2019ganai,selim2021ct,selim2021stan,selim2022cross,selim2022uda,roschewitz2024counterfactual}. However, there is limited understanding of how image dynamics influence deep learning–based downstream tasks -- an area that is both critical and often overlooked in the medical imaging domain. Recent approaches typically rely on CNNs to perform image translation and improve downstream performance across vendors without examining the underlying factors contributing to domain shifts. Moreover, these methods apply pixel-wise adjustments to the input, introducing the risk of hidden artifacts, shortcut learning, or even the removal of subtle diagnostic patterns due to their black-box nature.

These challenges motivate us to investigate how image dynamics influence a model’s generalization ability from a low-level image processing perspective. To this end, we apply several popular normalization methods to images from various vendors and observe varying degrees of performance degradation in the downstream task. Specifically, we find that X-ray images exhibit exposure ``errors'' similar to those in conventional photography, such as over-/underexposure and nonlinear sensor response. Building on these insights, we reframe the image harmonization task as an exposure correction problem and propose Global Deep Curve Estimation (GDCE), which learns a global nonlinear correction between source and reference scanners. Unlike black-box models, GDCE compensates for exposure differences using a pre-defined polynomial function without altering local patterns, thereby remaining transparent to users. Extensive experiment results on breast density classification using the EMBED dataset \cite{jeong2023emory} and pneumonia detection using the RSNA Pneumonia Detection Challenge dataset \cite{2018rsna} show that the proposed method effectively improves model generalization.

\section{Related work}
\label{sec:relatedwork}
In a typical transfer learning setup, a model is first pre-trained on a large-scale source dataset and then fine-tuned on a smaller, domain-specific target dataset. While ImageNet \cite{deng2009imagenet} is a widely used general-purpose source dataset, target datasets in medical imaging are often tied to specific acquisition pipelines. Fine-tuned models frequently exhibit poor generalization when confronted with acquisition pipeline shifts (which are the focus of this paper, although other shifts such as population shift may also occur). Since these acquisition shifts arise from variations in imaging pipelines across vendors, we review the literature in two related areas: medical image harmonization and image dynamics enhancement.

\subsection{Medical image harmonization}
Medical image harmonization aims to standardize image statistics. For X-ray images, this typically includes grayscale normalization, contrast enhancement, and denoising \cite{seoni2024all}. While hand-crafted methods have been explored \cite{deng2016mammogram,perre2018influence,cao2021breast,perez2020deep}, deep learning–based approaches have been the focus in recent years. These methods aim to translate images from various sources to a reference domain so that the resulting statistics resemble the target. Liang \emph{et al.} trained a generative adversarial network (GAN) to map images into a robust space in which the discriminator could not predict their source \cite{liang2019ganai}, with similar approaches proposed in \cite{selim2021ct,selim2021stan}. Selim \emph{et al.} employed two learning strategies for handling paired and unpaired data: paired data from the same vendor were used to train a standard GAN, while unpaired data across vendors were translated between domains via cycle consistency \cite{selim2022cross,selim2022uda}. This method was further extended by training a variational autoencoder to extract latent embeddings, which were then standardized using a score-based denoising diffusion probabilistic model \cite{selim2023latent}. A related work \cite{roschewitz2024counterfactual} used a deep structural causal model \cite{ribeiro2023high} to simulate realistic domain variations as data augmentation in counterfactual contrastive learning.

While these studies showed improved performance in downstream tasks, most applied pixel-wise domain translation, which carries the risk of introducing undesired artifacts and lacks transparency regarding how the images were normalized \cite{salahuddin2022transparency}. In contrast, we investigate how image dynamics relate to model generalization and propose enhancing images using a pre-defined polynomial function. Having full knowledge of the applied transformation helps prevent artifacts and promotes greater transparency in medical image analysis.

\subsection{Image dynamics enhancement}
Image dynamics enhancement aims to restore degraded dynamic ranges caused by overexposure, underexposure, or low-light conditions. It typically involves both local and global adjustments to enhance details and produce a more visually balanced tone. Eilertsen \emph{et al.} proposed a hybrid dynamic range autoencoder that directly converts a single Standard Dynamic Range (SDR) image into its High Dynamic Range (HDR) counterpart \cite{eilertsen2017hdr}. Endo \emph{et al.} introduced an indirect approach that predicts multiple SDR images at different exposure levels and merges them into a single HDR image \cite{endo2017deep}. Zhang \emph{et al.} developed a dual-illumination adjustment framework that performs conjugate exposure correction on the original and inverted images, respectively \cite{zhang2019dual}. Marnerides \emph{et al.} designed a multi-head CNN that learns local and global adjustments separately before fusing them into the final HDR output \cite{marnerides2018expandnet}. Liu \emph{et al.} reversed the image signal processing pipeline and compensated for compressed dynamics at each stage using two CNNs \cite{liu2020single}. Afifi \emph{et al.} simulated various exposure errors and used synthetic data to train a multi-scale enhancement model based on the Laplacian pyramid \cite{afifi2021learning}.

 In our experiments, we found that image dynamics, such as bit depth, display window, and sensor response curve, vary across vendors, resulting in both pronounced and subtle differences in visual appearance. Inspired by the studies discussed above, we reframe medical image normalization as an exposure correction problem and propose a method to adjust image dynamics accordingly. The proposed GDCE improves model generalization performance effectively without altering local structures.

\section{Methodology}
\label{sec:method}
We begin with a brief overview of the proposed method. Given multiple datasets obtained from different acquisition pipelines (\emph{e.g.}, scanners), we select one as the reference domain and treat the others as domain-shifted. A pathology classifier is first trained on the reference domain dataset and then kept fixed. This classifier is subsequently used as a ``domain discriminator'' to train the enhancement network, GDCE. Once training converges, GDCE is used to pre-process domain-shifted images before performing downstream tasks (\emph{e.g.}, pathology classification).

\subsection{Global deep curve estimation}
\begin{figure}[htb]
    \centering
    \includegraphics[scale=1]{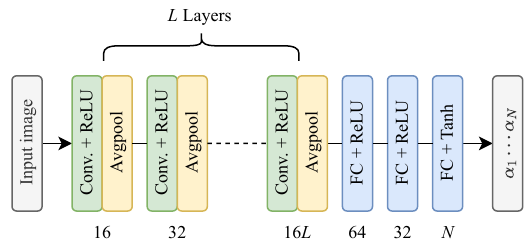}
    \caption{Block diagram of GDCE architecture. It consists of $L$ layers of ``convolution and average pooling'' to extract image features, followed by three fully connected layers to predict the coefficients $\alpha_{n}$ for Eq.~\ref{fig-dgce}.}
    \label{fig-dgce}
\end{figure}

Inspired by \cite{guo2020zero}, we design a CNN termed GDCE to perform global exposure correction. The model architecture is illustrated in Fig.~\ref{fig-dgce}. It consists of several convolution layers for input image feature extraction, followed by a multi-layer perceptron that maps the features to a series of coefficients $\alpha_{n}$, which are then applied to the image iteratively as follows:

\begin{equation}
     I_{n} = I_{n-1} + \alpha_{n}I_{n-1}\left ( 1 - I_{n-1} \right ),\;n=1,2,\dots,N-1,N,
     \label{eq-dce}
\end{equation}
where $n$ denotes the iteration index, $N$ is the total number of iterations, and $\alpha_{n}$ represents the predicted correction coefficient. This formulation has many benefits: First, it approximates the polynomial through iterative compositions of simpler functions, allowing the model to better utilize its learning capacity than directly fitting a high-order polynomial; Second, although GDCE is capable of capturing both global and local features, Eq.~\ref{eq-dce} defines exposure correction as a global operation, enabling transparent data manipulation without the risk of local artifects; Third, since $I_{0}\in \left [ 0,1 \right ]$ and $\alpha_{n}\in \left [ -1,1 \right ]$, the resulting image $I_{n}$ at each iteration remains within the range $\left [ 0,1 \right ]$, thereby avoiding data clamping that could potentially discard useful information.

Here, we emphasize the distinction between using the proposed GDCE as a pre-processing step before downstream tasks and fine-tuning the downstream models directly on new data, as in \cite{jimenez2023memory}. GDCE can be regarded as a specialized form of image normalization based on pre-defined enhancement operations, whereas fine-tuning adjusts the parameters of the downstream model to accommodate domain shifts.

\subsection{Task-oriented domain shift alignment}
Since the selected scanners are from different vendors, obtaining paired data across vendors is impractical. Consequently, many existing works \cite{liang2019ganai,selim2021ct,selim2021stan} adopt unsupervised learning approaches, particularly those based on GANs. Inspired by the use of discriminators in GANs, we propose a simple yet effective strategy to train GDCE by leveraging the downstream task. In our case, we use a pre-trained pathology classifier as the ``domain discriminator''. The domain classification loss $\mathscr{L}_{\text{CE}}$ is defined as:

\begin{equation}
     \mathscr{L}_{\text{CE}}(I_{N}, y_{\text{gt}}) = - \sum_{j=1}^{C} y_{\text{gt},j} \log \left( \frac{\exp\left( \mathcal{D}_j\left( I_{N} \right) \right)}{\sum_{k=1}^{C} \exp\left( \mathcal{D}_k\left( I_{N} \right) \right)} \right),
     \label{eq-ce-loss}
\end{equation}
where $\mathcal{D}$ denotes the ``domain discriminator'' (\emph{i.e.}, the downstream task model), $y_{\text{gt}}$ is the ground-truth label, $j$ indicates the class index, and $C$ is the total number of classes. Although we refer to the pathology classifier as a ``discriminator'', the training process is not adversarial.

In addition, we use a perceptual loss $\mathscr{L}_{\text{P}}$ to preserve the visual appearance of the enhanced image, defined as:

\begin{equation}
    \mathscr{L}_{\text{P}} = \left\| \mathcal{V}_{i} \left( I_{N} \right) - \mathcal{V}_{i} \left( I_{R} \right) \right\|_{1},
    \label{eq-peceptual-loss}
\end{equation}
where $I_{R}$ is an arbitrary image from the reference scanner, and $\mathcal{V}_{i}$ denotes the output of the $i$-th layer from a pre-trained VGG backbone. We use VGG-16 as the perceptual model \cite{johnson2016perceptual}, with $i = 7$ (\emph{i.e.}, ``conv. 2-2''), which is the shallowest layer that enables model convergence. This design choice ensures that the perceptual loss captures low-level visual characteristics while avoiding high-level semantic features.

The total loss $\mathscr{L}_{\text{T}}$ is defined as the summation of the classification and perceptual losses:

\begin{equation}
    \mathscr{L}_{\text{T}} = \mathscr{L}_{\text{P}} + \mathscr{L}_{\text{CE}}.
    \label{eq-total-loss}
\end{equation}

The intuition behind task-oriented domain shift alignment is similar to ``fine-tuning'' the downstream task model on new scanner data, except that GDCE is trained to enhance the input image in a way that improves downstream task performance. Although conceptually straightforward, this approach does not guarantee meaningful enhancement of the input data. Nguyen \emph{et al.} showed that when activations from the discriminator are back-propagated, subtle yet deterministic patterns can also be transferred to the input \cite{nguyen2016synthesizing}. In our context, this implies that the enhancer may unintentionally learn to embed hidden adversarial perturbations as shortcuts by manipulating local patterns \cite{xiao2018generating}.

To address these issues, we design GDCE to adjust only the global tone curve, without the capacity to modify local pixel-level details. This imposes a hard constraint that prevents the ``domain discriminator'' from back-propagating adversarial perturbations to the enhancer during training.

\subsection{Datasets and implementation}
We train a medical image classification model as a downstream task using transfer learning with a ResNet-50 backbone \cite{tajbakhsh2016convolutional}. As the source dataset, we adopt RadImageNet \cite{mei2022radimagenet} due to its slightly better performance \cite{juodelyte2024source} and robustness to noise \cite{lu2024exploring}. For the target datasets, we generate two small subsets: one from the EMBED dataset \cite{jeong2023emory} for breast density classification (\emph{i.e.}, classes A–D, evaluated based on the BI-RADS breast density scale \cite{sickles2013acr}), and the other from the RSNA Pneumonia Detection Challenge \cite{2018rsna} for pneumonia detection.

\begin{figure}[htb]
    \centering
    \input{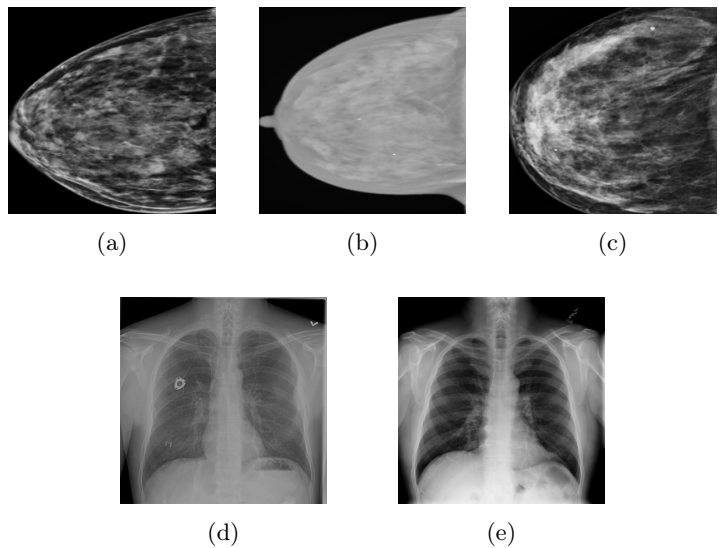}
    \caption{Visual examples of full-range images from the selected scanners for comparison. (a). Selenia (the reference scanner); (b). Senograph; (c). Clearview; (d). Brand X; (d). Brand Y (the reference scanner).}
    \label{fig-overview}
\end{figure}

To simplify the analysis, we select samples from three scanners for the breast density classification task: Clearview CSm, Selenia Dimensions (used as the reference scanner), and Senograph 2000D. For the pneumonia detection task, we select samples from two scanners with pixel spacings of 0.171 and 0.143, with the latter used as the reference scanner. Table~\ref{tbl-dataset} summarizes the training and test set partitions, and Fig.~\ref{fig-overview} presents full-range image examples from the selected scanners. \footnote{The training data from the reference scanner is used solely for fine-tuning the downstream task model and for evaluating GDCE predictions in terms of low-level perceptual similarity.}

\begin{table}[htb]
    \centering
    \caption{Summary of the train/test dataset partitions}
    \label{tbl-dataset}
    \setlength{\tabcolsep}{5pt}
    \renewcommand{\arraystretch}{1.2}
    \begin{tabular}{cccccc}
        \hline
        \multirow{2}{*}{Scanner model} & \multirow{2}{*}{Refer to} & \multicolumn{4}{c}{Breast density from \cite{jeong2023emory}}\\
        \cmidrule(lr){3-6} %
        & & A & B & C & D \\
        \hline
        Clearview CSm & Clearview & 0/123 & 400/100 & 400/100 & 0/81\\
        Selenia Dimensions & Selenia & 400/100 & 400/100 & 400/100 & 400/100\\
        Senograph 2000D & Senograph & 0/52 & 400/100 & 400/100 & 0/94\\
        \hline
        \hline
        \multirow{2}{*}{Scanner model} & \multirow{2}{*}{Refer to} & \multicolumn{4}{c}{Pneumonia from \cite{2018rsna}}\\
        \cmidrule(lr){3-6} %
        & & \multicolumn{2}{c}{Positive} & \multicolumn{2}{c}{Negative}\\
        \hline
        Pixel spacing 0.171 & Brand X & \multicolumn{2}{c}{400/100} & \multicolumn{2}{c}{400/100}\\
        Pixel spacing 0.143 & Brand Y & \multicolumn{2}{c}{400/100} & \multicolumn{2}{c}{400/100}\\
        \hline
    \end{tabular}
\end{table}

It is worth noting that there are no training samples for classes A and D in the Clearview and Senograph scanners due to an insufficient number of available cases. However, the absence of such training data can serve as a strong indicator of model overfitting, as the domain shift of interest is semantically independent of the labels. During training, images from the source and reference scanners are passed to GDCE, which is supervised using the loss defined in Eq.~\ref{eq-total-loss}. At each epoch, the model is validated only on the source scanner, and the performance is reported as the worst accuracy across groups (\emph{i.e.}, A–D for breast density and positive/negative for pneumonia), as this metric better captures the lower bound of model performance \cite{compton2023more}.

For breast density classification, we report the confusion matrix for per-class evaluation and the Receiver Operating Characteristic Area Under the Curve (ROC-AUC) score for overall performance. For pneumonia detection, we compute precision and recall as the evaluation metrics. All reported metrics represent the average results over 5-fold cross-validation.

The experiments were implemented using PyTorch with an image resolution of $512 \times 512$, a learning rate of $10^{-4}$, a batch size of 12, and the Adam optimizer. These hyper-parameters were empirically selected based on common practices from our previous projects.

\section{Experiment results}
\label{sec:experiment}
\subsection{Image dynamics affect model generalization}
We tested the breast density classifier using two image normalization methods: full-range normalization and display window clamping. Note that for the Clearview and Selenia scanners the two methods are identical, thus we omit them and only analyze performance change on the Senograph data.

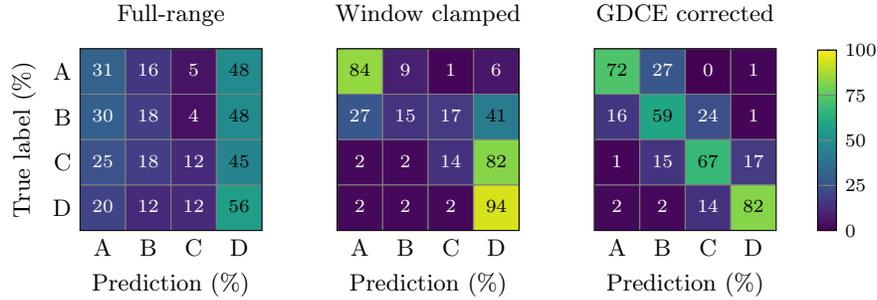
\begin{figure}[htb]
    \centering
    \begin{tikzpicture}
    \begin{groupplot}[
        group style={
            group name=confusion,
            group size=3 by 2,
            horizontal sep=0.55cm,
            vertical sep=1.1cm,
        },
        width=4.25cm, height=4.25cm,
        colormap name=viridis,
        xlabel=Prediction (\%),
        ylabel=True label (\%),
        xtick={0,...,3},
        xticklabels={A, B, C, D},
        xtick style={draw=none},
        ytick={0,...,3},
        yticklabels={A, B, C, D},
        ytick style={draw=none},
        enlargelimits=false,
        xticklabel style={rotate=0},
        colorbar=false,
        xlabel style={font=\normalsize},
        ylabel style={font=\normalsize},
        title style={font=\normalsize},
        ticklabel style={font=\normalsize},
        every axis title/.append style={yshift=-0.5ex},
        point meta=explicit,
        point meta min=0,
        point meta max=100,
        visualization depends on={\thisrow{C} \as \cellvalue},
        nodes near coords={
            \pgfmathparse{\cellvalue < 35 ? "white" : "black"}%
            \edef\temp{\noexpand\textcolor{\pgfmathresult}{\pgfmathprintnumber{\cellvalue}}}%
            \temp
        },
        nodes near coords style={
            yshift=-5pt,
            font=\normalsize,
        }
    ]

    \nextgroupplot[
        title={\strut Full-range},
        xlabel={},
        xticklabels={}
    ]
    \addplot[
        matrix plot,
        mesh/cols=4,
        point meta=explicit,
        draw=gray
    ] table [meta=C] {
        x y C
        0 0 4
        1 0 15
        2 0 1
        3 0 80
        0 1 6
        1 1 14
        2 1 0
        3 1 80
        0 2 11
        1 2 9
        2 2 0
        3 2 80
        0 3 17
        1 3 4
        2 3 1
        3 3 78
    };

    \nextgroupplot[
        title={\strut Window clamped},
        ylabel={},
        yticklabels={},
        xlabel={},
        xticklabels={}
    ]
    \addplot[
        matrix plot,
        mesh/cols=4,
        point meta=explicit,
        draw=gray
    ] table [meta=C] {
        x y C
        0 0 84
        1 0 9
        2 0 1
        3 0 6
        0 1 27
        1 1 15
        2 1 17
        3 1 41
        0 2 2
        1 2 2
        2 2 14
        3 2 82
        0 3 2
        1 3 2
        2 3 2
        3 3 94
    };

    \nextgroupplot[
        title={\strut z-Score normalized},
        ylabel={},
        yticklabels={},
        xlabel={},
        xticklabels={}
    ]
    \addplot[
        matrix plot,
        mesh/cols=4,
        point meta=explicit,
        draw=gray
    ] table [meta=C] {
        x y C
        0 0 37
        1 0 11
        2 0 10
        3 0 42
        0 1 47
        1 1 9
        2 1 9
        3 1 35
        0 2 52
        1 2 4
        2 2 5
        3 2 39
        0 3 42
        1 3 1
        2 3 2
        3 3 55
    };
    
    \nextgroupplot[
        title={\strut Stan-CT~\cite{selim2021stan}}
    ]
    \addplot[
        matrix plot,
        mesh/cols=4,
        point meta=explicit,
        draw=gray
    ] table [meta=C] {
        x y C
        0 0 20
        1 0 6
        2 0 18
        3 0 56
        0 1 20
        1 1 2
        2 1 19
        3 1 59
        0 2 18
        1 2 3
        2 2 20
        3 2 59
        0 3 19
        1 3 4
        2 3 19
        3 3 58
    };

    \nextgroupplot[
        title={\strut CVH-CT~\cite{selim2022cross}},
        ylabel={},
        yticklabels={},
    ]
    \addplot[
        matrix plot,
        mesh/cols=4,
        point meta=explicit,
        draw=gray
    ] table [meta=C] {
        x y C
        0 0 56
        1 0 31
        2 0 2
        3 0 11
        0 1 12
        1 1 29
        2 1 26
        3 1 33
        0 2 0
        1 2 3
        2 2 32
        3 2 65
        0 3 0
        1 3 0
        2 3 8
        3 3 92
    };

    \nextgroupplot[
        title={\strut GDCE},
        ylabel={},
        yticklabels={},
    ]
    \addplot[
        matrix plot,
        mesh/cols=4,
        point meta=explicit,
        draw=gray
    ] table [meta=C] {
        x y C
        0 0 72
        1 0 27
        2 0 0
        3 0 1
        0 1 16
        1 1 59
        2 1 24
        3 1 1
        0 2 1
        1 2 15
        2 2 67
        3 2 17
        0 3 2
        1 3 2
        2 3 14
        3 3 82
    };
    \end{groupplot}

    \begin{axis}[
        at={(confusion c3r2.south east)},
        xshift=0.25cm,
        yshift=0cm,
        hide axis,
        scale only axis,
        height=6.43cm,
        width=0cm,
        colormap name=viridis,
        colorbar,
        point meta min=0,
        point meta max=100,
        colorbar style={
        width=0.25cm, %
        ytick={0,25,50,75,100},
        yticklabel style={font=\normalsize},
        }
        ]
        \addplot [draw=none] coordinates {(0,0)};
    \end{axis}
\end{tikzpicture}
    \caption{Confusion matrices of the Senograph test set by six methods. The corresponding ROC-AUC are 0.52, 0.80, 0.58, 0.50, 0.82, and 0.89, respectively.}
    \label{fig-test-senograph}
\end{figure}

The results are shown in Fig.~\ref{fig-test-senograph}. We noticed that the performance of the Senograph scanner improved after switching to display window clamping, indicating that the clamped data had a lower level of domain shift. This was confirmed after we further inspected the visual appearance. As shown in Fig.~\ref{fig-visual-compare}, the original dynamic range of the breast in the Senograph image was low; in contrast, the image clamped by the display window exhibited much richer details.

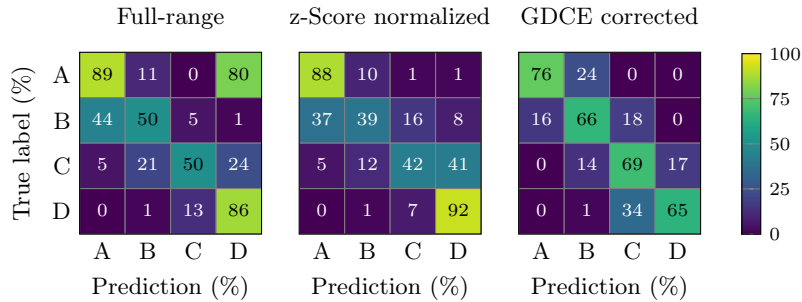
\begin{figure}[htb]
    \centering
    \begin{tikzpicture}
    \begin{groupplot}[
        group style={
            group name=confusion,
            group size=3 by 2,
            horizontal sep=0.55cm,
            vertical sep=1.1cm,
        },
        width=4.25cm, height=4.25cm,
        colormap name=viridis,
        xlabel=Prediction (\%),
        ylabel=True label (\%),
        xtick={0,...,3},
        xticklabels={A, B, C, D},
        xtick style={draw=none},
        ytick={0,...,3},
        yticklabels={A, B, C, D},
        ytick style={draw=none},
        enlargelimits=false,
        xticklabel style={rotate=0},
        colorbar=false,
        xlabel style={font=\normalsize},
        ylabel style={font=\normalsize},
        title style={font=\normalsize},
        ticklabel style={font=\normalsize},
        every axis title/.append style={yshift=-0.5ex},
        point meta=explicit,
        point meta min=0,
        point meta max=100,
        visualization depends on={\thisrow{C} \as \cellvalue},
        nodes near coords={
            \pgfmathparse{\cellvalue < 35 ? "white" : "black"}%
            \edef\temp{\noexpand\textcolor{\pgfmathresult}{\pgfmathprintnumber{\cellvalue}}}%
            \temp
        },
        nodes near coords style={
            yshift=-5pt,
            font=\normalsize,
        }
    ]

    \nextgroupplot[
        title={\strut Full-range},
        xlabel={},
        xticklabels={}
    ]
    \addplot[
        matrix plot,
        mesh/cols=4,
        point meta=explicit,
        draw=gray
    ] table [meta=C] {
        x y C
        0 0 85
        1 0 13
        2 0 0
        3 0 2
        0 1 49
        1 1 37
        2 1 8
        3 1 6
        0 2 9
        1 2 15
        2 2 39
        3 2 37
        0 3 0
        1 3 1
        2 3 10
        3 3 89
    };

    \nextgroupplot[
        title={\strut Window clamped},
        ylabel={},
        yticklabels={},
        xlabel={},
        xticklabels={}
    ]
    \addplot[
        matrix plot,
        mesh/cols=4,
        point meta=explicit,
        draw=gray
    ] table [meta=C] {
        x y C
        0 0 85
        1 0 13
        2 0 0
        3 0 2
        0 1 49
        1 1 37
        2 1 8
        3 1 6
        0 2 9
        1 2 15
        2 2 39
        3 2 37
        0 3 0
        1 3 1
        2 3 10
        3 3 89
    };

    \nextgroupplot[
        title={\strut z-Score normalized},
        ylabel={},
        yticklabels={},
        xlabel={},
        xticklabels={}
    ]
    \addplot[
        matrix plot,
        mesh/cols=4,
        point meta=explicit,
        draw=gray
    ] table [meta=C] {
        x y C
        0 0 88
        1 0 10
        2 0 1
        3 0 1
        0 1 37
        1 1 39
        2 1 16
        3 1 8
        0 2 5
        1 2 12
        2 2 42
        3 2 41
        0 3 0
        1 3 1
        2 3 7
        3 3 92
    };
    
    \nextgroupplot[
        title={\strut Stan-CT~\cite{selim2021stan}}
    ]
    \addplot[
        matrix plot,
        mesh/cols=4,
        point meta=explicit,
        draw=gray
    ] table [meta=C] {
        x y C
        0 0 21
        1 0 34
        2 0 23
        3 0 2
        0 1 22
        1 1 25
        2 1 25
        3 1 28
        0 2 24
        1 2 13
        2 2 20
        3 2 43
        0 3 26
        1 3 13
        2 3 12
        3 3 49
    };
    
    \nextgroupplot[
        title={\strut CVH-CT~\cite{selim2022cross}},
        ylabel={},
        yticklabels={},
    ]
    \addplot[
        matrix plot,
        mesh/cols=4,
        point meta=explicit,
        draw=gray
    ] table [meta=C] {
        x y C
        0 0 70
        1 0 29
        2 0 0
        3 0 1
        0 1 18
        1 1 61
        2 1 17
        3 1 4
        0 2 0
        1 2 12
        2 2 44
        3 2 44
        0 3 0
        1 3 2
        2 3 6
        3 3 92
    };

    \nextgroupplot[
        title={\strut GDCE},
        ylabel={},
        yticklabels={},
    ]
    \addplot[
        matrix plot,
        mesh/cols=4,
        point meta=explicit,
        draw=gray
    ] table [meta=C] {
        x y C
        0 0 63
        1 0 36
        2 0 1
        3 0 0
        0 1 12
        1 1 70
        2 1 18
        3 1 0
        0 2 1
        1 2 17
        2 2 62
        3 2 20
        0 3 0
        1 3 1
        2 3 28
        3 3 71
    };
    \end{groupplot}

    \begin{axis}[
        at={(confusion c3r2.south east)},
        xshift=0.25cm,
        yshift=0cm,
        hide axis,
        scale only axis,
        height=6.43cm,
        width=0cm,
        colormap name=viridis,
        colorbar,
        point meta min=0,
        point meta max=100,
        colorbar style={
        width=0.25cm, %
        ytick={0,25,50,75,100},
        yticklabel style={font=\normalsize},
        }
        ]
        \addplot [draw=none] coordinates {(0,0)};
    \end{axis}
\end{tikzpicture}
    \caption{Confusion matrices of the Clearview test set by siz methods. The corresponding ROC-AUC are 0.87, 0.87, 0.88, 0.51, 0.87, and 0.88, respectively.}
    \label{fig-confmat-clear}
\end{figure}

However, the performance improvement was limited, which we believe is due to two reasons. First, radiologists often manually adjust the display window to improve readability, introducing domain shift and model bias. Second, unlike computed tomography, where the data is calibrated to match Hounsfield units, X-ray images do not follow a universal standard. Consequently, the sensor's response and the corresponding gamma correction might not share the same linearity across vendors. This was further observed from the poor performance of the Clearview data in Fig.~\ref{fig-confmat-clear}. Although visually similar to Selenia data, the low-density tissues appeared brighter, while the high-density regions and the air pixels showed marginal differences.

\subsection{GDCE helps improve generalization performance}
We processed all the test images using the proposed GDCE and reran the pathology classifiers. For a broader comparison, we also trained and tested two GAN-based methods from the literature \cite{selim2021stan,selim2022cross}. In breast density classification, we observed performance improvements for the Clearview and Senograph scanners. Specifically, the overall ROC-AUC score for the Senograph scanner increased substantially to 0.89, the highest among all competitors. For the Clearview test set, labels B and C showed improvement after exposure correction by GDCE. Similar results in pneumonia classification, as shown in Table~\ref{tbl-rsna-result}, further supported its effectiveness in domain shift alignment.

\begin{table}[htb]
  \centering
  \caption{Results of pneumonia classification (Brand X)}
  \label{tbl-rsna-result}

  \begin{threeparttable}
    \setlength{\tabcolsep}{5pt}
    \renewcommand{\arraystretch}{1.2}

    \begin{tabular}{@{}cccccc@{}}
      \toprule
      & Baseline & z-Score & Stan-CT~\cite{selim2021stan} & CVH-CT~\cite{selim2022cross} & GDCE \\
      \midrule
      Precision & 54.55\% & 58.23\% & 51.15\% & 67.58\% & 74.54\% \\
      Recall    & 95.20\% & 96.40\% & 80.00\% & 76.40\% & 72.50\% \\
      \bottomrule
    \end{tabular}

    \begin{tablenotes}
      \scriptsize
      \item[*] Brand Y: 79.06\% (Precision) and 72.60\% (Recall).

    \end{tablenotes}
  \end{threeparttable}
\end{table}

\begin{figure}[htb]
    \centering
    \input{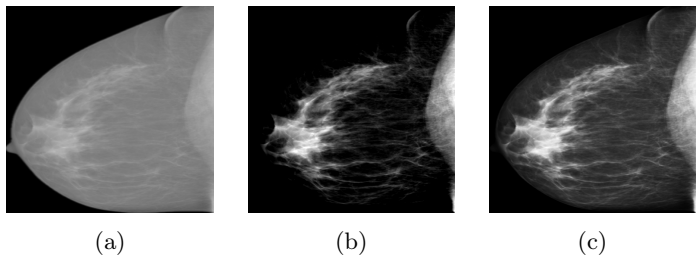}
    \caption{Visual comparison of the same Sanograph image handled by three methods. (a). Full-range normalized; (b). Display window clamped; (c). Exposure corrected by GDCE.}
    \label{fig-visual-compare}
\end{figure}

To further compare the processed data, we plot a sample image in Fig.~\ref{fig-visual-compare}. We observed that the processed images from the Senograph scanner perceptually looked quite similar to that of the Selenia scanner. These observations highlighted the merits of GDCE and demonstrated its effectiveness in improving the model's generalization ability across different target datasets.

\begin{figure}[htb]
    \centering
    \input{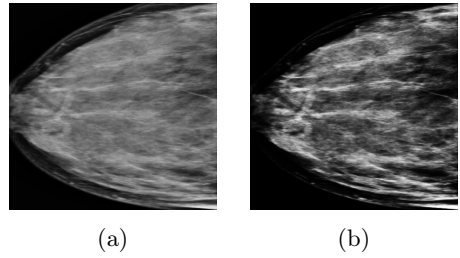}
    \caption{A failure case from the Clearview scanner. (a) Original image in full range; (b). Exposure corrected by GDCE.}
    \label{fig-failure}
\end{figure}

We also observed performance degradation in extreme cases (\emph{i.e.}, class A and D) from the Clearview scanner. By analyzing the failure cases (Fig.~\ref{fig-failure} shows an example), a possible explanation is that the lack of such data during GDCE training (see Table~\ref{tbl-dataset}) led the model to overly enhance the ``dark'' images (\emph{i.e.}, class A) while suppressing the ``bright'' images (\emph{i.e.}, class D), thereby limiting its ability to handle such outliers. In contrast, the dynamic range of the original Senograph images was quite low, which may have allowed GDCE to better extract exposure features from extreme cases, even when such samples were not present in training.

\subsection{Shallow model suffices for acquisition shift alignment}
\begin{figure}[htb]
    \centering
    \begin{tikzpicture}
    \begin{groupplot}[
        group style={
        group name=confusion,
        group size=2 by 1,
        horizontal sep=0.5cm,
        vertical sep=0.5cm,
        },
        width=6cm, height=4.5cm,
        colormap name=viridis,
        xlabel=Convolution layers,
        ylabel=Iterations,
        xtick={0,...,10},
        xticklabels={2,3,4,5,6,7,8,9,10,11,12},
        xtick style={draw=none},
        ytick={0,...,6},
        yticklabels={8,7,6,5,4,3,2},
        ytick style={draw=none},
        enlargelimits=false,
        xticklabel style={rotate=0},
        colorbar=false,
        xlabel style={font=\normalsize},
        ylabel style={font=\normalsize},
        title style={font=\normalsize},
        ticklabel style={font=\normalsize},
        every axis title/.append style={yshift=-0.5ex},
        point meta=explicit,
        point meta min=30,
        point meta max=80,
    ]

    \nextgroupplot[
        title={\strut Validation set},
    ]
    \addplot[
        matrix plot,
        mesh/cols=11,
        point meta=explicit,
        draw=gray
    ] table [meta=C] {
        x y C
        0 0 72
        1 0 75
        2 0 79
        3 0 76
        4 0 74
        5 0 78
        6 0 75
        7 0 74
        8 0 75
        9 0 72
        10 0 74
        0 1 64
        1 1 75
        2 1 72
        3 1 76
        4 1 76
        5 1 74
        6 1 76
        7 1 79
        8 1 78
        9 1 75
        10 1 80
        0 2 66
        1 2 78
        2 2 72
        3 2 78
        4 2 76
        5 2 75
        6 2 75
        7 2 78
        8 2 75
        9 2 80
        10 2 75
        0 3 74
        1 3 74
        2 3 75
        3 3 74
        4 3 74
        5 3 76
        6 3 76
        7 3 74
        8 3 75
        9 3 75
        10 3 75
        0 4 69
        1 4 74
        2 4 74
        3 4 74
        4 4 74
        5 4 74
        6 4 74
        7 4 75
        8 4 74
        9 4 74
        10 4 75
        0 5 57
        1 5 70
        2 5 70
        3 5 54
        4 5 69
        5 5 74
        6 5 70
        7 5 74
        8 5 59
        9 5 57
        10 5 59
        0 6 59
        1 6 55
        2 6 55
        3 6 55
        4 6 55
        5 6 56
        6 6 55
        7 6 55
        8 6 56
        9 6 55
        10 6 55
    };

    \nextgroupplot[
        title={\strut Test set},
        ylabel={},
        yticklabels={},
    ]
    \addplot[
        matrix plot,
        mesh/cols=11,
        point meta=explicit,
        draw=gray
    ] table [meta=C] {
        x y C
        0 0 59
        1 0 52
        2 0 60
        3 0 53
        4 0 40
        5 0 46
        6 0 43
        7 0 31
        8 0 30
        9 0 36
        10 0 30
        0 1 66
        1 1 57
        2 1 58
        3 1 44
        4 1 44
        5 1 37
        6 1 41
        7 1 43
        8 1 38
        9 1 37
        10 1 51
        0 2 69
        1 2 58
        2 2 60
        3 2 52
        4 2 53
        5 2 33
        6 2 41
        7 2 43
        8 2 36
        9 2 52
        10 2 26
        0 3 68
        1 3 59
        2 3 41
        3 3 65
        4 3 36
        5 3 53
        6 3 42
        7 3 33
        8 3 27
        9 3 26
        10 3 38
        0 4 67
        1 4 58
        2 4 51
        3 4 43
        4 4 52
        5 4 41
        6 4 52
        7 4 47
        8 4 30
        9 4 33
        10 4 27
        0 5 63
        1 5 62
        2 5 69
        3 5 59
        4 5 62
        5 5 51
        6 5 59
        7 5 56
        8 5 62
        9 5 62
        10 5 63
        0 6 62
        1 6 62
        2 6 58
        3 6 62
        4 6 58
        5 6 63
        6 6 61
        7 6 62
        8 6 61
        9 6 62
        10 6 62
    };

    \end{groupplot}

    \begin{axis}[
        at={(confusion c2r1.south east)},
        anchor=south west,
        xshift=0.25cm,
        yshift=0cm,
        hide axis,
        scale only axis,
        height=2.91cm,
        width=0cm,
        colormap name=viridis,
        colorbar,
        point meta min=30,
        point meta max=80,
        colorbar style={
            width=0.25cm,
            ytick={30,55,80},
            yticklabel style={font=\normalsize},
        }
    ]
    \addplot [draw=none] coordinates {(0,0)};
    \end{axis}
\end{tikzpicture}
    \caption{Worst group performance by convolution layers and iterations on the Clearview validation and test sets.}
    \label{fig-abla}
\end{figure}
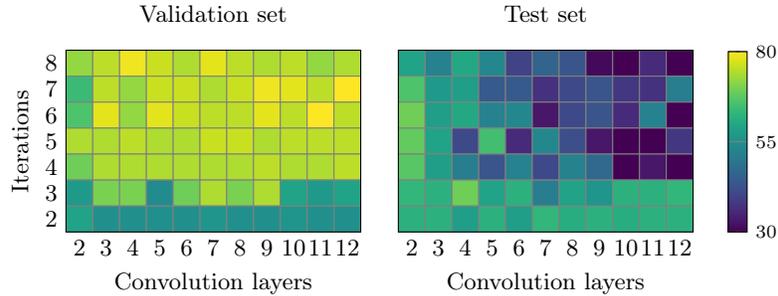

As the only enhancement we applied was the global tone curve adjustment, in theory the resulting GDCE should have been simple enough without requiring attention to high-level fine patterns. To verify this, we performed an ablation study that compared the performance of different model configurations. Specifically, we focused on two factors: the number of convolution layers in Fig.\ref{fig-dgce}, and the number of iterations in Eq.\ref{eq-dce}. The results~\footnote{We found that the worst-performing group in the training sets consistently achieved around 99\%, regardless of the number of iterations or convolutional layers.} on the validation set and the test set are presented as heatmaps in Fig.~\ref{fig-abla}.

At first glance, the two heatmaps appeared to be opposed: test performance dropped as validation performance rose. This was expected, as breast density classes A and D were not present in the training set. A validation performance higher than that of the classifier on the reference scanner indicated that GDCE was overfitting to the training data, which became more apparent with increased learning capacity (\emph{i.e.}, more convolution layers). Interestingly, even for a shallow GDCE with 7 convolution layers, the overfitting issue became significant. We therefore questioned the need to employ much deeper, pixel-wise image processing models to compensate for acquisition domain shifts, such as those presented in Section~\ref{sec:relatedwork}.

\section{Discussion and conclusions}
\label{sec:concl}
\subsection{Significance of results}
Our experiments showed the effectiveness of the proposed GDCE over commonly used image normalization methods such as z-score normalization and display window clamping in medical image analysis, supporting the idea that global image dynamics is a key factor in acquisition shifts. Since the enhancement is performed via a pre-defined polynomial function, our method offers a more transparent understanding of how the image is processed. Moreover, the ablation study showed that a shallow GDCE is sufficient to compensate for image dynamic differences, raising concerns about the necessity of deeper, pixel-wise architectures in the existing literature.

\subsection{Limitations and future directions}
Despite its effectiveness, this work has several limitations. First, our ablation studies did not allow us to further investigate how the absence of certain labels in the training data might influence label-independent image dynamics; Second, we did not consider other forms of acquisition shifts that could also bias the classifier (\emph{e.g.}, samples of breast density class D might be noisier due to higher X-ray absorption in high-density tissues \cite{lu2024exploring}). These represent potential directions for future research.

\subsection{Concluding remarks}
In this paper, we observed that acquisition dynamics influence a model’s generalization ability across different vendors in X-ray medical image analysis. Specifically, we found that scanner dynamics are not only determined by the radiologist’s preferred display window but are also related to the sensor’s nonlinear response. We therefore conclude that X-ray imaging can suffer from exposure-related errors too. To address this, we proposed GDCE, which corrects such errors iteratively using a pre-defined polynomial function. Experiment results showed improved test performance on three scanners that were unseen during downstream task training. Future work could involve a more detailed analysis of model overfitting and other forms of acquisition shifts to further improve generalization ability.

\begin{credits}
\subsubsection{\ackname} This research was supported by Novo Nordisk Foundation under Grant NNF21OC0068816.

\subsubsection{\discintname}
The authors have no competing interests to declare that are relevant to the content of this article.
\end{credits}

\bibliographystyle{splncs04}
\bibliography{refs-yucl}

\end{document}